\documentclass{article}

\usepackage{PRIMEarxiv}

\usepackage[utf8]{inputenc} % allow utf-8 input
\usepackage[T1]{fontenc}    % use 8-bit T1 fonts
\usepackage{hyperref}       % hyperlinks
\usepackage{url}            % simple URL typesetting
\usepackage{booktabs}       % professional-quality tables
\usepackage{amsfonts}       % blackboard math symbols
\usepackage{nicefrac}       % compact symbols for 1/2, etc.
\usepackage{microtype}      % microtypography
\usepackage{lipsum}
\usepackage{fancyhdr}       % header
\usepackage{graphicx}       % graphics
\graphicspath{{media/}}     
\usepackage{tikz}
\usepackage{comment}
\usepackage{amsmath,amssymb}  
\usepackage{color}
\usepackage{multirow}
\usepackage{pifont} 
\newcommand{\cmark}{\ding{51}}%
\newcommand{\xmark}{\ding{55}}%
\usepackage{tabularx}

\usepackage{subcaption}
\usepackage{floatrow}
 
\newfloatcommand{capbtabbox}{table}[][\FBwidth]
\usepackage{wrapfig}
 
\title{Temporal Contrastive Learning with Curriculum}

\author{
  Shuvendu Roy, 
    Ali Etemad \\
  Dept. ECE and Ingenuity Labs Research Institute \\
Queen's University, Kingston, Canada\\
  \texttt{\{shuvendu.roy, ali.etemad\}@queensu.ca} 
}

\begin{document}
\maketitle

\begin{abstract}
We present ConCur, a contrastive video representation learning method that uses curriculum learning to impose a dynamic sampling strategy in contrastive training. More specifically, ConCur starts the contrastive training with easy positive samples (temporally close and semantically similar clips), and as the training progresses, it increases the temporal span effectively sampling hard positives (temporally away and semantically dissimilar). To learn better context-aware representations, we also propose an auxiliary task of predicting the temporal distance between a positive pair of clips. We conduct extensive experiments on two popular action recognition datasets, UCF101 and HMDB51, on which our proposed method achieves state-of-the-art performance on two benchmark tasks of video action recognition and video retrieval. We explore the impact of encoder backbones and pre-training strategies by using R(2+1)D and C3D encoders and pre-training on Kinetics-400 and Kinetics-200 datasets. Moreover, a detailed ablation study shows the effectiveness of each of the components of our proposed method.
\end{abstract}

% keywords can be removed

\section{Introduction}\label{sec:introduction}
Self-supervised learning (SSL) has seen tremendous growth in recent years \cite{chen2020simple,jing2020self} for learning image representations, and has achieved state-of-the-art results in a variety of different downstream applications \cite{chen2020simple,roy2021self}. Since SSL can learn important and discriminative representations from input data without any human-annotated labels, it greatly reduces the need for large-scale supervised datasets. The most primitive class of self-supervised learning learns the data representations by defining a pre-text objective on the unlabeled data, for example, predicting the type (or amount) of transformations applied on the input. Various pre-text recognition tasks have been since proposed and shown to be effective for different SSL tasks \cite{srivastava2015unsupervised,wang2019self,gan2018geometry}. Recently a class of self-supervised methods called contrastive learning \cite{chen2020simple,he2020momentum} has shown significant improvement and generalization capabilities across different tasks and domains. The basic idea of contrastive learning is to distinguish between positive and negative samples, where the positives are different views of the same input image (usually generated by augmentations), while the negatives are derived from different inputs.

The success of contrastive learning for video representations has shown a similar trend to that of image representation learning. A number of prior works have directly adopted the popular image-based contrastive methods and applied them to videos with the aid of an additional sampling step for the clips \cite{feichtenhofer2021large,qian2021spatiotemporal}. Unlike image contrastive learning that only applies augmentations to generate two positive samples, videos consist of a temporal dimension from which different sub-clips are sampled and defined as either positive \cite{qian2021spatiotemporal} or negatives \cite{sermanet2018time}. The sampling technique of clips and the definition of positive samples is still an open problem for video contrastive learning as different solutions resort to different strategies for this purpose \cite{feichtenhofer2021large,qian2021spatiotemporal}. For example, in \cite{feichtenhofer2021large}, sub-clips were sampled from the entire video and any pair was treated as positive samples. However, it was argued in \cite{qian2021spatiotemporal} that frames that are temporally far apart contain totally different contextual information, and thus considering them as positive pairs would be unreasonable. Therefore for deriving the positive pairs, they used a sampling technique that assigned a probability value inversely proportional to the distance between the frames. On the other extreme, distanced clips in the same video were used as negatives in \cite{sermanet2018time}. This issue motivates our paper where we pose the following problems. (\textbf{1}) How should the positive pairs be defined in contrastive video representation learning? (\textbf{2}) Can positive temporal pairs be selected dynamically (from a temporal perspective) without using a pre-defined definition?

\begin{figure}[t]
    \centering
    \includegraphics[width=0.75\textwidth]{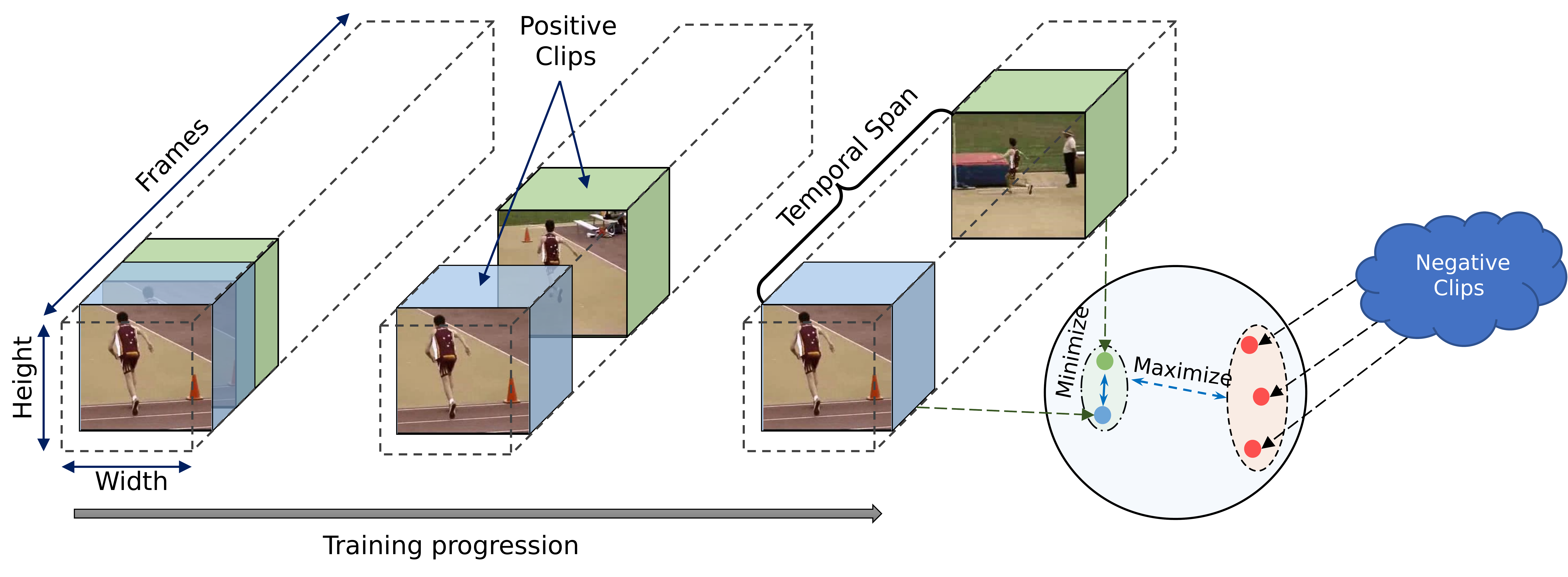}
    \caption{An overview of ConCur. 
    } 
    \label{fig:overview}
\end{figure}

To tackle these problems, we present a method for \textbf{Con}trastive learning using context-aware \textbf{Cur}riculum learning (ConCur), a self-supervised approach to learn video representations. Our method samples multi-instance positive pairs from a dynamic temporal span and progressively increases the range, in essence gradually \textit{hardening} the positive samples. More specifically, at the beginning of the training, ConCur samples positive clips that are temporally overlapped, thus containing similar semantic contexts. As the training progresses we increase the temporal span from which positive samples are randomly sampled, effectively increasing the probability of sampling positives that are temporally far apart and semantically dissimilar. We use a modified Multi-instance MoCo (MI-MoCo) \cite{he2020momentum} as the contrastive loss for our proposed method. In order to learn better temporal representations, we also propose a \textit{Context Similarity} loss term that facilitates learning of context-aware video representations by predicting the temporal frame distance between any two positive clips. We rigorously evaluate ConCur on widely used public datasets for two downstream tasks, namely activity recognition and video retrieval, and show that our method achieves state-of-the-art results when using different popular video encoders (e.g., R(2+1)D \cite{tran2018closer}, C3D \cite{tran2015learning}). An overview of ConCur is presented in Fig. \ref{fig:overview}.

We make the following contributions in this paper.
 
    (\textbf{1}) We introduce a temporal sampling strategy for sampling positive clips with a constraint on the dynamic temporal span. We show that randomly sampling positives over the entire duration of a video hurts the representation learnt by contrastive loss. 
    % \item 
    (\textbf{2}) We propose a curriculum learning strategy for contrastive video learning, that increases the temporal span from which positive clips are randomly sampled, effectively increasing the harness for the contrastive loss over a given period. The proposed curriculum learning module increases the accuracy of the downstream application with no increase in model size or FLOPs.
    % \item 
    (\textbf{3}) We also propose a \textit{Context Similarity} loss term to learn better temporal representation by predicting the temporal distance between the two positive clips. 
    % \item 
    (\textbf{4}) We conduct extensive experiments on two popular action recognition datasets, UCF101 and HMDB51, and set a new state-of-the-art result for self-supervised video action recognition and video retrieval task.

\section{Related Works}\label{sec:related_works}
In this section, we discuss the literature in three key areas related to this paper: self-supervised video representation learning with pre-text tasks, self-supervised contrastive video representation learning, and curriculum learning. 
 
\noindent \textbf{Self-supervised Video Learning with Pre-text Tasks.}
A variety of different pre-text tasks have been proposed in the literature for learning video representations in a self-supervised manner. A temporal cycle consistency loss was proposed in \cite{dwibedi2019temporal} with a pre-text task of temporal alignment. To learn coherent features, a future frame prediction task was proposed in \cite{srivastava2015unsupervised}. Motion and appearance features were used in \cite{wang2019self}, while frame color estimation was used as a pre-text task in \cite{vondrick2018tracking}. Two other popular pre-text tasks that worked on frame-level information include video pace prediction \cite{wang2020self} and video frame sorting \cite{xu2019self,lee2017unsupervised,kim2019self,fernando2017self}. Another similar loss was proposed in \cite{yang2020video}, where clips sampled with different sampling rates were used to learn consistent representations. For more accurate video recognition, four different transformations were utilized in \cite{jenni2020video}, namely speed change, random sampling, periodic change, and frame warp. 

Multi-modality was utilized in different prior work \cite{wang2019learning,korbar2018cooperative,dwibedi2019temporal,gan2018geometry}, where information from one modality was has been used as a self-supervisory signal for other modalities. For example, the flow and video were used in \cite{gan2018geometry} to learn better representations, while several works have utilized audio with video \cite{wang2019learning,korbar2018cooperative,dwibedi2019temporal}. 
     
\noindent \textbf{Self-supervised Contrastive Video Learning.} 
In recent days, there have been a great amount of progress in image representation learning using contrastive learning \cite{chen2020simple,he2020momentum}. Video representation learning has also seen a similar trend \cite{feichtenhofer2021large,qian2021spatiotemporal}. Some recent works have utilized direct temporal modification of image-based methods \cite{feichtenhofer2021large,qian2021spatiotemporal,roy2021spatiotemporal}, i.e., by introducing a frame-sampling step. However, in video contrastive learning settings, sampling video clips to form positive and negative pairs are one of the key components of video contrastive learning methods. In \cite{sermanet2018time}, a contrastive loss was proposed to attract different viewpoints of the same input clip, while repealing the clip from other samples of the same video that are far apart. 

Some prior works have proposed to use all clips from the same video as positive samples \cite{feichtenhofer2021large}. 
% Similarly, to allow all clips of a video to be considered positives, 
To do so, a technique was introduced in \cite{qian2021spatiotemporal} that sampled the clips with a probability that was inverse to the distance. 
In \cite{chen2021rspnet}, a contrastive loss was utilized with clips of different speeds to learn video feature representations. Some works also adopted hybrid approaches \cite{pan2021videomoco} by incorporating additional loss terms to the contrastive framework. For example, a momentum contrastive learning approach was used in \cite{pan2021videomoco} with adversarial learning. Finally, following the success of masked prediction in natural language processing, a masked prediction loss was utilized with contrastive learning in \cite{tan2021vimpac}.

% \subsection{Curriculum Learning}
\noindent \textbf{Curriculum Learning.} Curriculum learning is a paradigm that is inspired by the human learning behaviour of staring with `easy' concepts and slowly learning more complex topics. This method was popularised in machine learning by \cite{bengio2009curriculum}. In classical machine learning, all the training data are presented to the model without a particular strategy. To incorporate curriculum learning, the model is initially presented with easy samples and later hard samples are slowly introduced in training \cite{bengio2009curriculum}. 

Broadly, curriculum learning methods are categorized into two groups: manually pre-defined difficulty measures \cite{tudor2016hard,wei2016stc,platanios2019competence} and automatic curriculum learning \cite{meng2017theoretical,weinshall2018curriculum}. While pre-defined difficulty measure based approaches utilize domain knowledge to control the progression of curriculum learning, automatic curriculum learning is generally a dynamic and domain-agnostic approach \cite{meng2017theoretical,weinshall2018curriculum}. Such methods have been common and effective in image \cite{tudor2016hard,wei2016stc} and text \cite{platanios2019competence} representation learning tasks. A detailed review on the topic can be accessed in \cite{wang2021survey}. Despite the high potential for using curriculum learning to dynamically adapt the sampling strategy in contrastive video representation learning, this concept has not yet been explored.

\section{Proposed Method}\label{sec:method}
In this section, we describe the components of our proposed method. First, we present the preliminaries for our approach, followed by the proposed ConCur method. 

\subsection{Preliminaries}
\textbf{Data.} Let $X=\{x_i\}, i \in [1,N]$ denote the training dataset, where $N$ is the total number of training video instances. Each video is a stack of RGB frames of size $3 \times T \times S^2$, where $T$ is the total number of frames, and $S$ is the spatial dimension of each frame. First, we re-sample the input video at a desired frame per second (fps) rate $f$. We then sample a clip from the video with $t$ consecutive frames to obtain a $3 \times t \times s^2$ clip, where $s$ is the input spatial dimensions to the model. 

\noindent \textbf{Data augmentation module.} A stochastic data augmentation module $\tau$ is performed on each clip $\{x_i\}, i \in [1, N_B]$ of a mini-batch, where $N_B$ is the batch size. Two random augmentations $t\sim\tau,\ t'\sim\tau$ are applied on the input clip $x_i$ to generate two corresponding views $(v_i, v_i')$. In contrastive learning settings, $(v_i, v_i')$ are considered a positive pair, whereas $(v_i, v_j')$, $(v_j, v_i')$ are considered negative pairs for $j \in [1, N_B], j\neq i$.
    
\noindent \textbf{Feature encoding.} A 3D CNN model (e.g., R(2+1)D or C3D), with a projection head (described below) is used as a feature extractor to generate the query embedding, $q = f_\theta (v_i)$. A momentum encoder generates the positive key embedding, $k^+=f_{\theta_m}(v_i')$. A dictionary stores the keys from the previous iterations of training, which are used as negatives $k^-$. There is no gradient update on the momentum encoder as it is updated by
\begin{equation}
    \theta_m = m\theta_m + (1-m)\theta ,
\end{equation}
where $m$ is the value for the momentum.

\noindent \textbf{Projection head.} Following common practice in self-supervised literature \cite{chen2020simple,he2020momentum}, we use a small multi-layer linear network as a projection head to linearly transform the output of the encoder to a different latent representation. This is proven to be useful for learning better representations with contrastive pre-training. More details about the instantiating are described in Section \ref{sub:imp_details}.
    
\noindent \textbf{Contrastive loss.} The aim of the contrastive loss is to learn from positive and negative samples by bringing the embedding of positive samples closer and pushing the negatives apart from the positives. Given an encoder query $q$, positive key $k^+$, and negative keys $k^-=\{k_0, k_1, ...\}$ from dictionary queue, we utilize a momentum contrastive loss named MoCo \cite{he2020momentum}. The MoCo loss function can be written as:
\begin{equation}\label{eq:nce}
    L_q=-\operatorname{log}\frac{\operatorname{exp}(sim(q, k^+)/\alpha)}{\sum_{k\in\{K^-\}}\operatorname{exp}(sim(q, k)/\alpha)} ,
\end{equation}
where $\alpha$ is a temperature parameter and $sim(q, k)$ is the cosine similarity represented as $sim(q,k)=(q^Tk)/(||q||~||k||)$. This loss is built upon the InfoNCE loss \cite{van2018representation} popularized by SimCLR \cite{chen2020simple}. The InfoNCE loss in its original setup utilized one positive sample (from the augmented view) and $N_B - 1$ negative samples. The InfoNCE loss was expanded as Multi-Instance InfoNCE \cite{miech2020end} to take multiple positives, where the total number of positive is denoted by $\rho$. Accordingly, $\rho$ transformations are sampled from $\tau$, and applied to the input image $x_i$ to generate $\{x_i^1, x_i^2, ... x_i^\rho\}$. A multi-instance setting was also explored in the context of videos with MoCo \cite{feichtenhofer2021large}, where different sub-clips sampled from the same video were considered positives.

\begin{figure}[t]
    \centering
    \includegraphics[width=1\columnwidth]{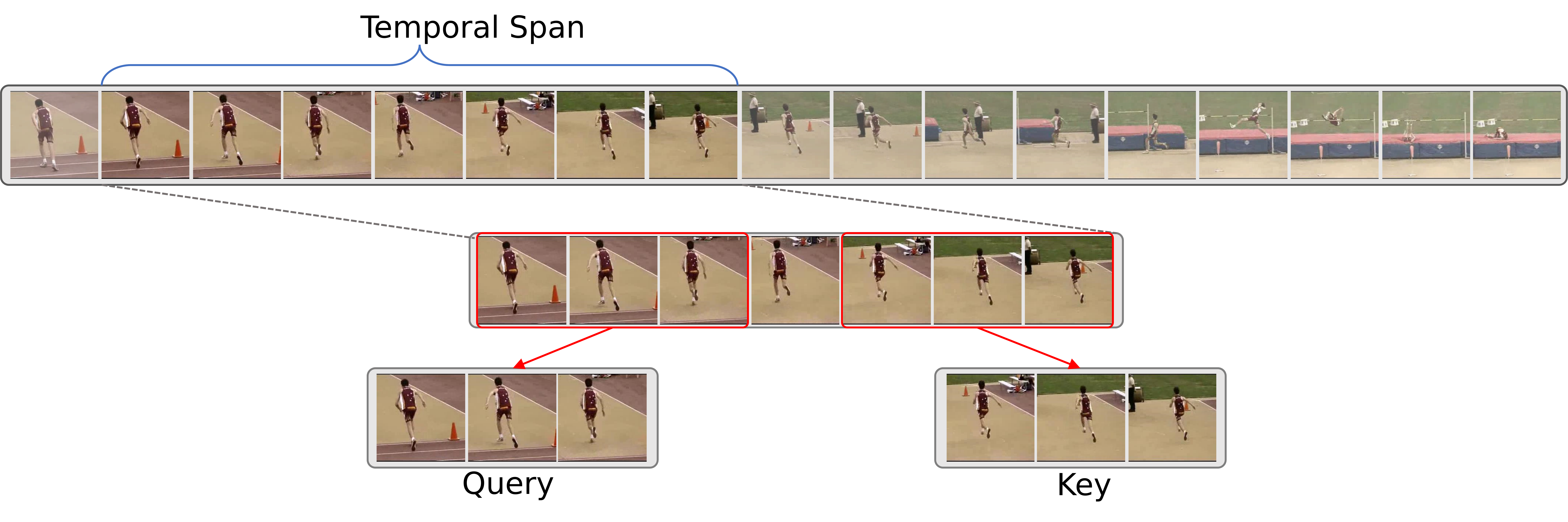}
    \caption{Illustration of our clip sampling. A frame span is selected on a long video to select a sub-video, from which $\rho$ clips are randomly picked.}
    \label{fig:sampling}
\end{figure}
    
\subsection{Multi-instance Sampling with Dynamic Frame Span}\label{sec:mi_moco} 
For a reasonably long video, the semantic context at the start of a video can be very dissimilar to the end of that video. This is why sampling positive clips over the entire duration of a video and blindly encouraging their respective embeddings to be situated close to one another is not reasonable. 
To tackle this problem, we propose a sampling method that imposes a dynamic constraint on the temporal span from which positive clips are sampled.
The proposed sampling technique only picks positive samples from inside the defined temporal span, and any clip sampled from a different video is considered a negative. A visual illustration of the proposed sampling technique is shown in Fig. \ref{fig:sampling}. Given, a video clip with a temporal length (number of frames) $T$, and a temporal span ($TS$), the method first picks a random starting frame $s\in[1, T-TS]$. Accordingly, the temporal window for positives is defined as $w\in [s, s+TS]$. We then sample $\rho$ clips of $t$ consecutive frames from the sampling window $w$, and $\rho$ random spatial augmentations $\{t^1, t^2, ... , t^\rho\}$ from $\tau$ which are applied on the sampled clips to get $\rho$ positive samples $x^+ = \{x_i^1, x_i^2, ... x_i^\rho\}$. In the multi-instance momentum contrastive setting, one positive sample is treated as the query and $\rho -1$ positive samples as keys. Following \cite{chen2020simple}, we adopt a symmetric version of the loss, where each clip in $x^+$ is considered as query and $\rho-1$ clips are considered positives $\{k^+\}$. The modified loss for Multi-instance MoCo is represented as follows (the total loss is averaged over $\rho$):
\begin{equation}\label{eq:mi_moco}
    L_{MI}=-\operatorname{log}\frac{{\sum_{k\in\{k^+\}}}\operatorname{exp}(sim(q, k^+)/\alpha)}{\sum_{k\in\{k^+, k^-\}}\operatorname{exp}(sim(q, k)/\alpha)}.
\end{equation}
In the following sub-section we describe how we use curriculum learning to progressively update $TS$.

\subsection{Curriculum Learning for Temporal Span Update}\label{sec:ccl}
\begin{figure}[t]
    \centering
    \includegraphics[width=\columnwidth]{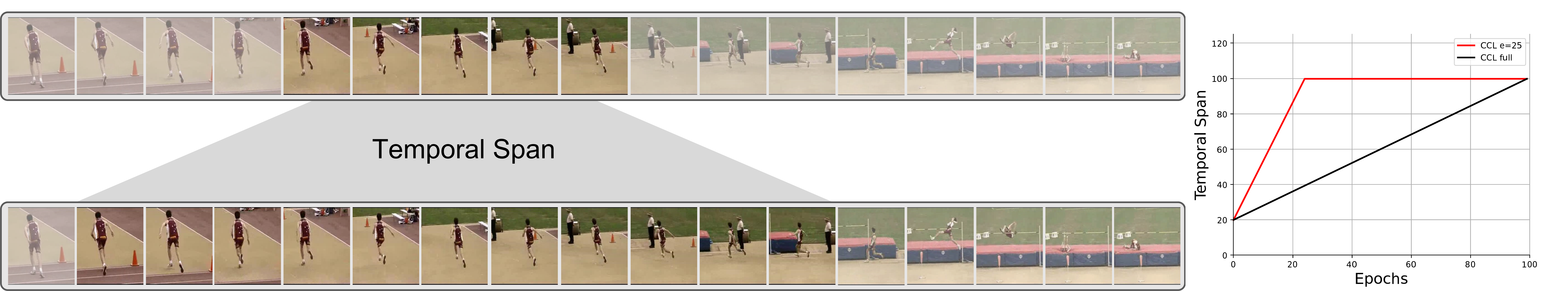}
    \caption{Illustration of curriculum learning on frame sampling, where the frame span for sampling positive increases over time.}
    \label{fig:ccl}
\end{figure}

In video contrastive settings, if a pair of sampled clips $(x_i^a, x_i^b)$ are very close or overlap with one another, they are more likely to contain semantically similar content. We define pairs that are \textit{temporally close} as `easy' positive pairs. On the other hand, pairs $(x_i^a, x_i^c)$ that are temporally far apart are considered `hard' positives. Here, we propose a hardness measure that gradually increases the temporal span, $TS$, of positive samples over the training epochs, effectively hardening the positive samples for our contrastive loss. The proposed curriculum learning component is illustrated in Fig. \ref{fig:ccl} (left), where the training starts with a short temporal span and is then increased over the training iterations. We investigate the proposed component in two settings. In first setting, we increase the hardness over the entire training phase. In the second setting, we limit the number of epochs over which hardening occurs ($E_{CL}$), and a constant hardness is used beyond that threshold (see Fig. \ref{fig:ccl} (right)). The temporal span at a given epoch $e$ is formulated as:
\begin{equation}
   TS_e = min\left(TS_m, ~TS_i + \left( \frac{TS_m - TS_i}{E_{CL}}\right) \times e \right),
\end{equation}
where $TS_m$ is the maximum temporal span, $TS_i$ is the initial temporal span, and $E_{CL}$ is the total number of epochs over which hardening is performed. 

\subsection{Context Similarity}\label{sec:pre_text}
We define an auxiliary task of predicting the temporal distance (number of frames) between any two positive clips (key and query) given their learned embeddings. This is done to help the model learn better contextual information regarding the location of the positive samples. To predict the distance between a query embedding $q$ and a key embedding $k$, we add a single linear layer that takes the concatenation of $q$ and $k$ as input and generates the context similarity prediction.

\begin{figure}
    \centering
    \includegraphics[width=0.75\columnwidth]{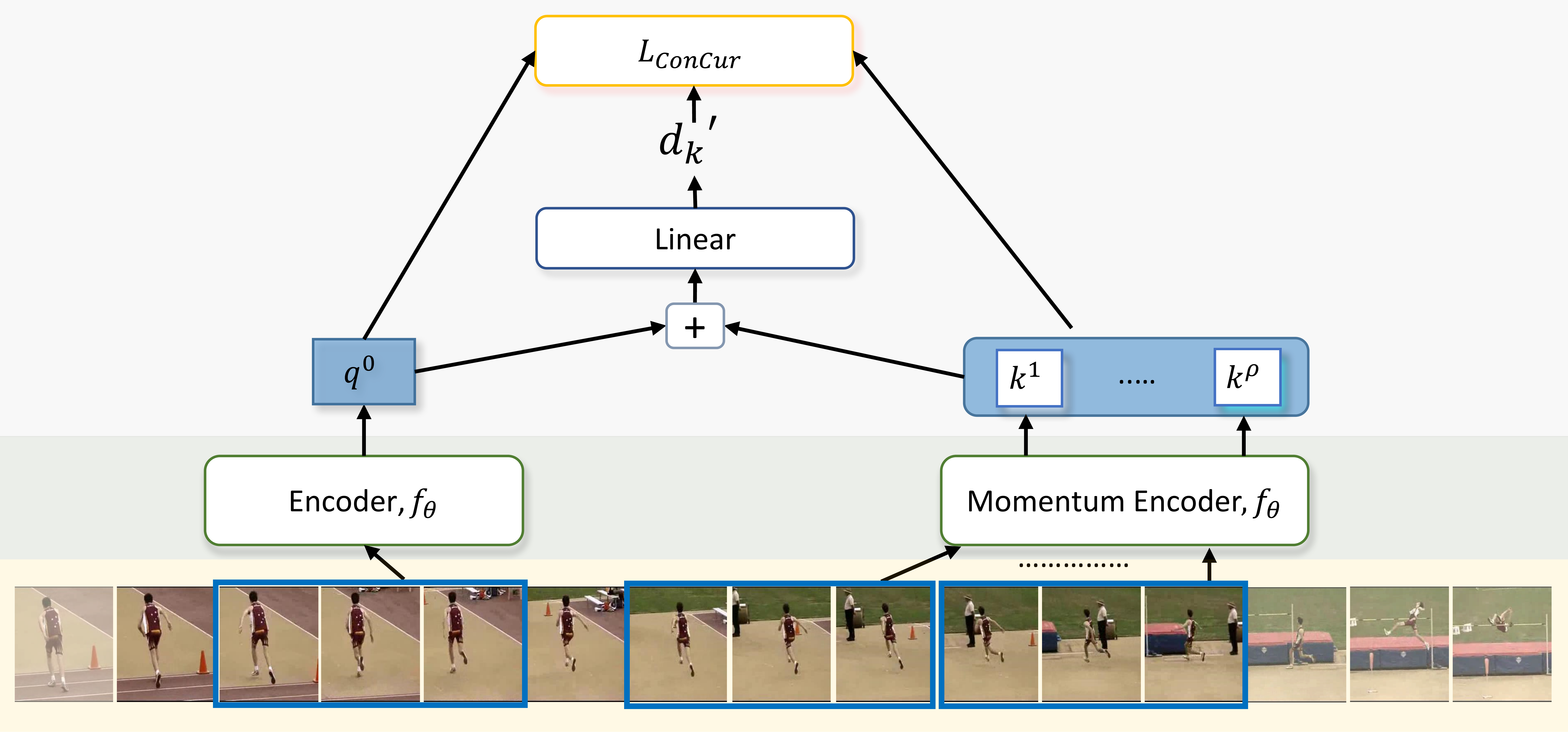}
    \caption{Visual illustration of the proposed model.}
  \label{fig:model}
\end{figure}

Let $\phi(x)$ be a function that returns the frame number at which a positive clip ($x$) starts within the original video.
We define the distance between the query clip $x_i^q$ and any positive key clip $x_i^k$ as $d_k = abs(\phi(x_i^q) - \phi(x_i^k))$, and the predicted distance by the model as $d_k'$. For $\rho$ positive keys in the proposed setting, we define the context similarity loss as: 
\begin{equation}
    L_{CS} = \sum_{k\in\{k^+\}} (d_k-d_k')^2 / \rho.
\end{equation}
    
Accordingly, the overall loss for ConCur is defined as:
\begin{equation}
    L_{ConCur} = L_{MI} + L_{CS},
\end{equation}
where $L_{MI}$ is the multi-instance MoCo loss presented earlier in Eq. \ref{eq:mi_moco}.  The overall diagram for our method is illustrated in Fig. \ref{fig:model}.

\begin{wraptable}{r}{8cm}
    \centering
    \setlength{\tabcolsep}{5pt}
        \begin{tabular}{c | c | c} 
            \hline
             Curriculum & CS loss & Accuracy (\%)\\
            \hline 
            % MoCo (curr.) &  80.64\\
            % MoCo + CS &  \\
            % MoCo (curr.) + CS & ??? \\ \hline
            \cmark & \cmark & \textbf{81.08} \\
            \cmark & \xmark &  80.64\\
            \xmark & \cmark &  80.17\\
            \xmark & \xmark & 79.46 \\
    
            % MI + CL + CS & \textbf{81.08}\%\\
            \hline
        \end{tabular}
        \caption{Ablation results.}
        \label{tab:abl}
\end{wraptable} 

\section{Experiments and Results}\label{sec:results}

In this section, we present the implementation details, experimental setup, and results of the proposed ConCur method. We perform extensive sensitivity and ablation studies, and compare our method with the state-of-the-art video self-supervised methods on two popular benchmark tasks of action recognition and video retrieval on two popular datasets namely UCF101 and HMDB51. The code for our method will be released publicly with the final version of the paper.

\subsection{Datasets}
Following other works in the area \cite{feichtenhofer2021large,pan2021videomoco}, we pre-train our model with Kinetics400 and Kinetics200, and fine-tune on UCF101 and HMDB51 datasets. In the following, we present a short description of each of the datasets used in this paper.

\noindent \textbf{Kinetics-400 (K400)} \cite{kay2017kinetics} is a large-scale dataset for video action recognition. It contains a total of about 306K videos. Most of the clips in this dataset have a length of 10 seconds. The videos are categorized into 400 action classes. The dataset contains training, test, and validation splits, where the training set contains close to 240K videos. We only use the training split in the pre-training stage of the method.
    
\noindent \textbf{Kinetics-200 (K200)} \cite{k200} is a subset of the K400 dataset. This subset contains 400 videos per class for 200 action classes, which gives a total of 80K videos for training. The validation set contains 25 examples per class. We use this dataset only for the ablation and sensitivity studies since this is a reasonable size dataset that requires less time to train the model. 
    
\noindent \textbf{UCF101} \cite{soomro2012ucf101} is a very popular action recognition dataset that contains over 13K videos collected from the Internet across 101 action classes. The dataset accumulates about 27 hours of video. It contains three training and testing splits. We report the average accuracy over the three splits. 

\noindent \textbf{HMDB51} \cite{kuehne2011hmdb} is another popular dataset for action recognition which contains around 6.7K videos with 51 action categories. This dataset is also divided into three training and testing splits. Since this is a comparatively smaller dataset, it is only used for fine-tuning and evaluation.

\subsection{Implementation Details}\label{sub:imp_details}
In this sub-section, we describe the detailed settings of our training along with other implementation details about contrastive pre-training of the model, downstream fine-tuning, and the evaluation protocol. We also summarize the video retrieval settings and the linear evaluation protocol.

First, we pre-train the video encoder with the proposed ConCur loss on Kinetics400. We report the performance of the method with two popular video encoders namely R(2+1d) \cite{tran2018closer} and C3D \cite{tran2015learning}. While all the ablation and sensitivity studies in this paper are conducted using the R(2+1)D encoder, other encoders are used in the final results and comparisons to the state-of-the-art. 
Following the self-supervised learning literature \cite{chen2020simple,he2020momentum}, we use an MLP projection head after the last layer (global pooling layer) of the video encoder, which embeds the output of the video representation into a vector of dimension 128. The MLP layer used here is a simple 2-layer fully connected network of sizes $(n \times 2048)$ and $(2048 \times 128)$. Here, $n$ is the output vector dimension of the video encoder. We use a ReLU activation layer between the fully-connected layers. The output of the MLP projection head is normalized before using the proposed loss function. Note that the MLP projection head is used only in the pre-training stage and is not involved in fine-tuning. Instead, a single linear layer is added on top of the output of the video encoder with output dimensions equal to the number of classes for the final prediction task.

The model is implemented with PyTorch framework and trained with 8 Nvidia V100 GPUs. The pre-training is done with a batch-size of 16 clips per GPU. Momentum SGD is used for the training with a momentum value of 0.9. Following \cite{chen2020simple}, a temperature value $\alpha$ of 0.07 is used. Shuffling BatchNorm \cite{he2020momentum} is utilized for the training. A base learning rate of 0.02 is used for training which is linearly warmed up from 0 to 0.02 over 5 epochs, and a cosine learning rate decay is used to reduce the learning rate for the rest of the training epochs. A weight decay regularization is used with a value of 0.0001. We pre-train the randomly initialized encoder for a total of 400 epochs for the final model. For all the ablation studies the encoder is trained for 100 epochs. Unless mentioned otherwise, we use $t=16$ as the number of frames with a spatial resolution of $s=112$ for the input to the model. 
The clips are selected after the original videos are re-sampled to 15 fps. For the momentum encoder, we use the key dictionary of size 65536, and a momentum value of 0.999.

For the data augmentation, we largely follow the augmentation protocol introduced in \cite{feichtenhofer2021large}. We follow Inception-style \cite{szegedy2015going} cropping with a minimum crop area of 0.2 and a maximum crop area of 0.76 following \cite{feichtenhofer2021large}. We apply random horizontal flip, color distortion, and Gaussian blur following SimCLR \cite{chen2020simple} and MoCo v2 \cite{chen2020improved}. The random gray scaling is applied with a probability of 0.2 and random horizontal flip with probability of 0.5. Following \cite{feichtenhofer2021large} we apply color distortion with a probability of 0.8 and color strength of $\{$\textit{brightness}, \textit{contrast}, \textit{saturation}, \textit{hue}$\} = {0.4s, 0.4s, 0.4s, 0.1s}$, with a default value of $s=0.5$. The Gaussian blur is applied with a probability of 0.5 using a spatial kernel with a standard dev. $\in$ [0.1, 2.0].

\begin{table*}[t!]
     
    \centering
    \setlength{\tabcolsep}{6.5pt}
    \resizebox{0.9\textwidth}{!}
    {
    \begin{tabular}{cccccccc} 
    \hline
        \multirow{2}{*}{Method} & \multirow{2}{*}{Backbone} & \multirow{2}{*}{Pre-train} & \multirow{2}{*}{Fine-tune} & \multirow{2}{*}{Res.} & \ \multirow{2}{*}{Frames} & \multicolumn{2}{c}{Datasets (\%)} \\
         &  & & & & & UCF-101 & HMDB-51 \\
        \hline
        CBT \cite{sun2019learning} & S3D & K600 & \xmark & 224 & 30 & 54.0 & 29.5 \\
        CCL \cite{kong2020cycle} & R3D-18 & K400 & \xmark & 112 & 8 & 52.1 & 27.8 \\
        MemDPC \cite{han2020memory}   & R3D-34 & K400 & \xmark & 224 & 40 & 54.1 & 27.8 \\
        TaCo \cite{bai2020can}  & R3D & K400& \xmark  & 112 & 16 & 59.6 & 26.7 \\
        RTT \cite{jenni2020video}   & C3D & UCF-101& \xmark & 112 & 16 & 60.6 & - \\
        MFO \cite{qian2021enhancing}  & S3D & K400& \xmark & 112 & 16  & 61.1 & 31.7 \\
        MFO \cite{qian2021enhancing}   & R3D-18 & K400 & \xmark & 112 & 16 & \underline{63.2} & \underline{33.4} \\
        \textbf{Ours} & R(2+1)D & K400  & \xmark & 112& 16& \textbf{67.4}& \textbf{45.2}\\
        
        \hline

        RTT \cite{jenni2020video} & C3D & UCF-101& \cmark & 112 & 16  & 68.3 & 38.4 \\
        RTT \cite{jenni2020video} & C3D & K400 & \cmark& 112 & 16  & 69.9 & 39.6\\
        PRP \cite{yao2020video}   & C3D & UCF-101& \cmark & 112 & 16  & 69.1 & 34.5 \\
        VCP \cite{luo2020video}   & C3D & UCF-101 & \cmark& 112 & 16  & 68.5 & 32.5 \\
        VCOP \cite{xu2019self}  & C3D & UCF-101 & \cmark & 112 & 16 & 65.6 & 28.4 \\
        Var.PSP \cite{cho2020self}   & C3D & UCF-101 & \cmark& 112 & 16  & 70.4 & 34.3 \\
        MoCo+BE \cite{wang2021removing}   & C3D & UCF-101 & \cmark & 112 & 16 & \underline{72.4} & \underline{42.3} \\
        RSPNet \cite{chen2021rspnet}  & C3D & K400 &\cmark & 112 & 16 &  \underline{76.7} & \underline{44.6} \\
        \textbf{Ours} & C3D & UCF-101 & \cmark & 112& 16 & \textbf{72.9}& \textbf{43.0}\\
        \textbf{Ours}  & C3D & K400  & \cmark& 112 & 16 &\textbf{77.9} & \textbf{48.2} \\
        \hline 
        
        VCP \cite{luo2020video} & R(2+1)D & UCF-101 & \cmark& 112 & 16  & 66.3 & 32.2 \\
        PRP \cite{yao2020video}   & R(2+1)D & UCF-101 & \cmark& 112 & 16  & 72.1 & 35.0 \\
        VCOP \cite{xu2019self}   & R(2+1)D & UCF-101 & \cmark & 112 & 16 & 72.4 & 30.9 \\
        Var.PSP \cite{cho2020self}  & R(2+1)D & UCF-101 & \cmark & 112 & 16 & 74.8 & 36.8 \\
        PacePred \cite{wang2020self}   & R(2+1)D & UCF-101& \cmark  & 112 & 16 & 75.9 & 35.9 \\
        RSPNet \cite{chen2021rspnet}   & R(2+1)D & K400  & \cmark& 112 & 16 & 81.1 & 44.6 \\
        RTT \cite{jenni2020video}  & R(2+1)D & UCF-101  & \cmark& 112 & 16 & \textbf{81.6} & \underline{46.4} \\
        VideoMoCo \cite{pan2021videomoco}  & R(2+1)D & K400 & \cmark  & 112 & 16 & \underline{78.7} & \underline{49.2} \\
        \textbf{Ours} & R(2+1)D & UCF-101 & \cmark  & 112 & 16 & \underline{78.1} & \textbf{49.5} \\
        \textbf{Ours} & R(2+1)D & K400 & \cmark  & 112 & 16 & \textbf{84.2} & \textbf{58.2} \\
         
    \hline
    \end{tabular}
    }
    \caption{Comparison with prior work on UCF-101 and HMDB-51 datasets. }   
    \label{tab:sota}
\end{table*}
    
\subsection{Evaluation Protocols}    
 
Following the standard video evaluation protocol \cite{tran2018closer,luo2020video,xu2019self}, we uniformly sample 10 clips from the test video alone its temporal axis. We then resize the smaller dimension to 112. Next, we take 3 spatial crops per clip (total 30 clips) with resolution $112 \times 112$ to cover full spatial space. Final prediction for a video is the average over all the clips.     

\noindent \textbf{Downstream fine-tuning.}
Following the pre-training step, we fine-tune the model with the pre-trained encoder on UCF101 and HMDB51 datasets. In this stage, we remove the MLP projection head and add a randomly initialized linear classification layer. The full model is then fine-tuned for 100 epochs. We fine-tune with an initial learning rate of 0.05. A cosine learning rate decay is utilized to reduce the learning over the epochs. A momentum SGD with momentum of 0.9 is used as optimizer. The model is trained with a batch size of 24 clips per GPU. Weight decay of 0.0001 is used as regularizer. Video clips are sampled at 15 fps and 16 consecutive frames of resolution $112\times112$ is used. For augmentation, we only use the random horizontal flip and random resize with the same values as the pre-training. 
    
\noindent \textbf{Linear evaluation.}
We also report the performance of the model on linear evaluation. Here, the pre-trained model is not fine-tuned in the fine-tuning stage; rather, only the randomly initialized linear classification layer is trained. The projection MLP head is removed before adding the linear layer. The rest of the training settings are kept the same as the full fine-tune protocol.

\noindent \textbf{Video retrieval.}
Following the evaluation protocol of \cite{luo2020video,xu2019self}, we also evaluate the proposed method on video retrieval. For this task, we pre-train the model on the UCF101 dataset, and use the pre-trained encoder (without projection head) to generate embeddings for all the videos in the training set. These training video embeddings are considered as keys. We then generate the embedding for the videos in the validation set, which are considered as queries. For each video in the test set (query), the top-K neighbours (keys) are recalled by calculating the cosine distances. When the original class label of the input query appears in the top-K keys, the prediction is considered as correct. This video retrieval evaluation is performed on UCF101 and HMDB51 dataset with different numbers of neighbours $K=1,5,10,20,50$.

\subsection{Curriculum Parameters}
Here, we study the key parameters of our curriculum learning module. Note that these studies were done on the smaller K200 dataset and trained for 100 epochs. 

\begin{figure}%
    \centering
    {{\includegraphics[width=5.5cm]{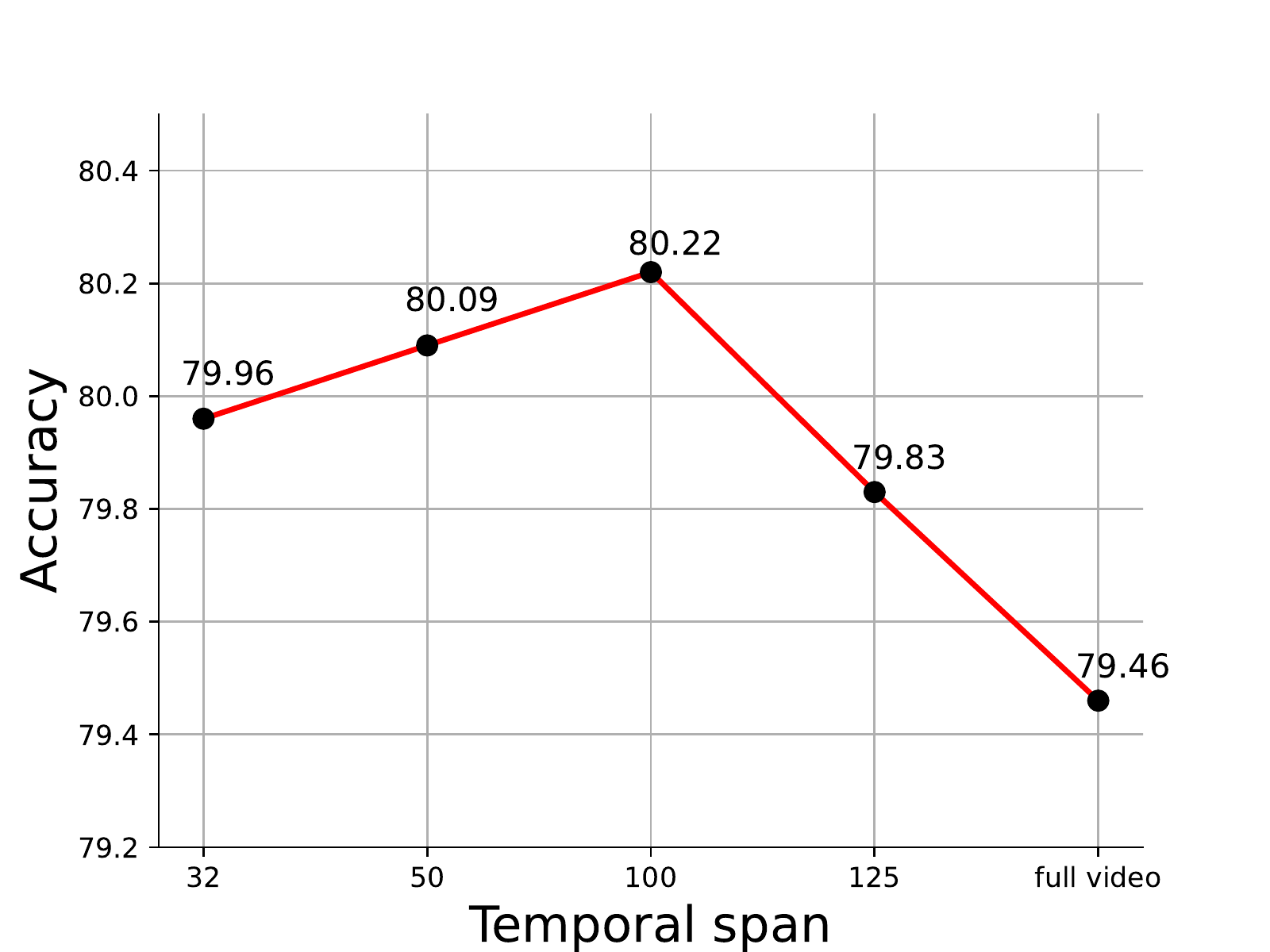} \label{fig:t_span}}}%
    \qquad
    {{\includegraphics[width=5.5cm]{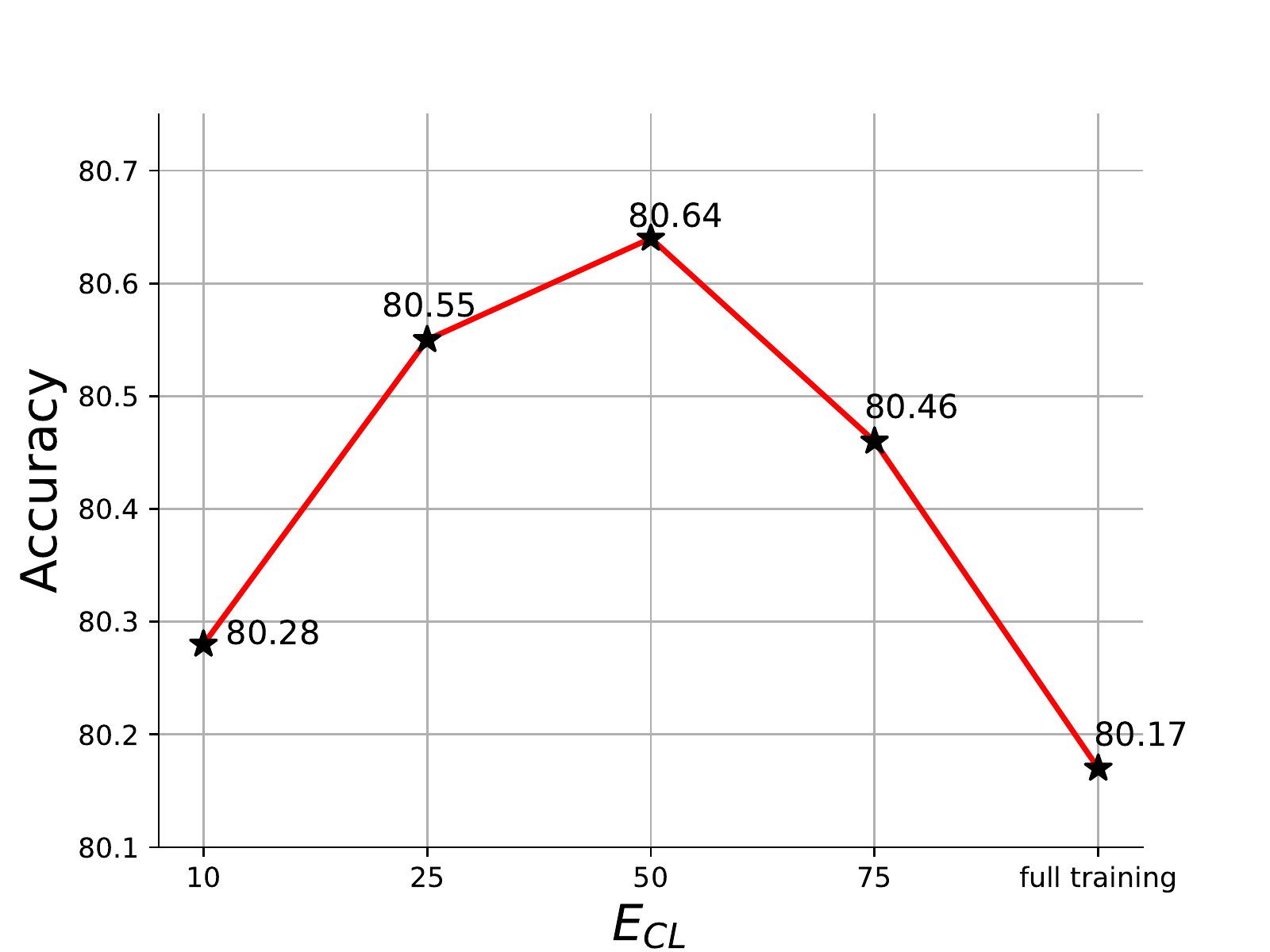}  \label{fig:e_cl} }}%
    \caption{Experiments on curriculum learning, where different temporal span values (top) and different values of $E_{CL}$ (bottom) are explored.}%
    % \vspace{-6pt}
    \label{fig:sensitibity}%
\end{figure}

First, we conduct an experiment for different temporal spans for sampling the positives. The results are presented in Fig. \ref{fig:sensitibity} (left), where it can be seen that the performance of the model increases when we increase the temporal span from 32 to 100. However, the accuracy drops when we use larger values. For example, a temporal span of 125 performs worse than 32. The accuracy is the lowest when the length of the full video is used as the temporal span. The reason behind this is likely the significant semantic contextual difference between the first and last parts of a long video. For the rest of the experiments, we set $TS = 100$.

\begin{table*}[t!]
    % \footnotesize
    \centering
    \setlength{\tabcolsep}{2.5pt}
    \resizebox{0.85\textwidth}{!}
    {
	\begin{tabular}{cc|ccccc|ccccc}
        \multirow{2}{*}{Method}  & \multirow{2}{*}{Backbone} & \multicolumn{5}{c}{UCF-101} & \multicolumn{5}{c}{HMDB-51} \\
         & & R@1 & R@5 & R@10 & R@20 & R@50 & R@1 & R@5 & R@10 & R@20 & R@50 \\
         \hline
         
        %  \hline
         VCOP \cite{xu2019self} &  C3D & 12.5 & 29.0 & 39.0 & 50.6 & 66.9 & 7.4 & 22.6 & 34.4 & 48.5 & 70.1 \\
         VCP \cite{luo2020video} & C3D & 17.3 & 31.5 & 42.0 & 52.6 & 67.7 & 7.8 & 23.8 & 35.3 & 49.3 & 71.6 \\
         PRP \cite{yao2020video} & C3D & 23.2 & 38.1 & 46.0 & 55.7 & 68.4 & 10.5 & 27.2 & 40.4 & 56.2 & 75.9 \\
         PacePred \cite{wang2020self} & C3D & 31.9 & 49.7 & 59.2 & 68.9 & 80.2 & 12.5 & 32.2 & 45.4 & 61.0 & 80.7 \\
         MoCo+BE \cite{wang2021removing} & C3D & - & - & - & - & - & 10.2 & 27.6 & 40.5 & 56.2 & 76.6 \\
         \textbf{Ours} & C3D & \textbf{32.2 }& \textbf{49.9}& \textbf{59.9}& \textbf{69.2} & \textbf{81.4} & \textbf{11.5} & \textbf{28.9} & \textbf{42.6} & \textbf{58.6} & \textbf{79.1}\\
         
        \hline
         VCOP \cite{xu2019self} & R(2+1)D & 10.7 & 25.9 & 35.4 & 47.3 & 63.9 & 5.7 & 19.5 & 30.7 & 45.6 & 67.0 \\
         VCP \cite{luo2020video} & R(2+1)D & 19.9 & 33.7 & 42.0 & 50.5 & 64.4 & 6.7 & 21.3 & 32.7 & 49.2 & 73.3 \\
         PRP \cite{yao2020video} & R(2+1)D & 20.3 & 34.0 & 41.9 & 51.7 & 64.2 & 8.2 & 25.3 & 36.2 & 51.0 & 73.0 \\
         PacePred \cite{wang2020self} & R(2+1)D & 25.6 & 42.7 & 51.3 & 61.3 & 74.0 & 12.9 & 31.6 & 43.2 & 58.0 & 77.1 \\
         \textbf{Ours} & R(2+1)D & \textbf{26.0 }& \textbf{42.9}& \textbf{52.7}& \textbf{61.4} & \textbf{83.8} & \textbf{13.0} & \textbf{32.1} & \textbf{43.4} & \textbf{60.0} & \textbf{80.1}\\
         
% 		\hline
		\hline
	\end{tabular}
    }
    \caption{Recall at top-K. Comparison with prior work for video retrieval task on UCF-101 and HMDB-51.}
    \label{tab:retrieval}
\end{table*}

Next, we experiment with different $E_{CL}$ values to identify its impact and optimal value. We present the results in Fig. \ref{fig:sensitibity} (right) where we observe that $E_{CL} = 50$ gives the best results, followed closely by $25$ and $75$. However, curriculum learning over a very short period (e.g., $E_{CL} = 10$) did not result in any notable improvements compared to not using curriculum learning. Moreover, large $E_{CL}$ values also do not help the model. This is in line with our expectation of identifying an optimum level of hardness, where harder or simpler self-supervisory signals hurt the performance \cite{appalaraju2020towards,sarkar2020self}.

\subsection{Ablation Experiments}
Here, we perform ablation experiments on K200 to investigate the effect of the key contributions of our work, namely \textit{curriculum learning} and \textit{Context Similarity} loss. The results are presented in Table \ref{tab:abl}.

As we observe from the table, removing the CS loss from ConCur reduces the accuracy by around 0.6\%, which shows the importance of the context similarity loss in the proposed method. It should be noted that in recent years, improvements made by the state-of-the-art are in a similar range, often improving prior works by approximately 0.5\% to 2.0\%. Next we remove the curriculum learning component, effectively allowing the positives to be sampled from the entire duration of the video, which shows a drop of 0.9\% accuracy on the downstream task. Finally, we train the model by removing both CL loss and curriculum learning and observe a drop of 1.6\% accuracy. This shows the effectiveness of each of the components of our proposed method.

\subsection{Comparison with State-of-the-art}
    
We present the main results in comparison with state-of-the-art methods on two video-related tasks, action recognition and video retrieval, which we perform on UCF101 and HMDB datasets as discussed earlier. 
    
\noindent \textbf{Action recognition.}
We present the results in Table \ref{tab:sota}. Two main encoder backbones, R(2+1)D and C3D encoders are used in the literature as well as our method. As discussed earlier, both linear evaluation and full fine-tuning are used in the experiments. When performing linear evaluation, we observe a huge improvement (12\%) on HMDB51 and a 4\% improvement on UCF-101. This is despite the fact that some of the related works in the table use larger models, large resolution inputs, or larger datasets to pre-train. 

As shown in Table \ref{tab:sota}, when we use the C3D encoder in ConCur and pre-train it on UCF101, we achieve SOTA for both UCF101 and HMDB51 datasets in the full fine-tuning scheme. When pre-trained on the larger K400 dataset, our model achieves 1.2\% and 3.6\% improvements relative to the previous SOTA results on UCF101 and HMDB51 respectively. When we use R(2+1)D pre-trained on UCF101, ConCur outperforms other methods on HMDB51, but not on UCF101, as we achieve the second-best results behind RTT \cite{jenni2020video}. However, ConCur achieves SOTA results on both of the datasets when R(2+1)D encoder is pre-trained on the larger K400 dataset. On HMDB51, we obtain a considerable improvement of 9\% in accuracy, and a 2.6\% improvement on UCF101.

\noindent \textbf{Video retrieval.}
The video retrieval results are presented in Table \ref{tab:retrieval}. Here, $\text{R@K}$ represents the recall at $K$ neighbours. From the table, we observe that ConCur achieves SOTA results using the C3D encoder on both UCF101 and HMBD51 datasets. We also observe good improvement when using the R(2+1)D encoder for different values of $K$ on both UCF101 and HMDB51 datasets. It can be noticed that the improvement of ConCur increases as $K$ is increased. For example, there is a 9.8\% improvement for UCF101 with $K=50$, and 3\% on HMBD51.

\subsection{Qualitative Analysis}
\begin{figure}[t]
    \centering
    \includegraphics[width=1\columnwidth]{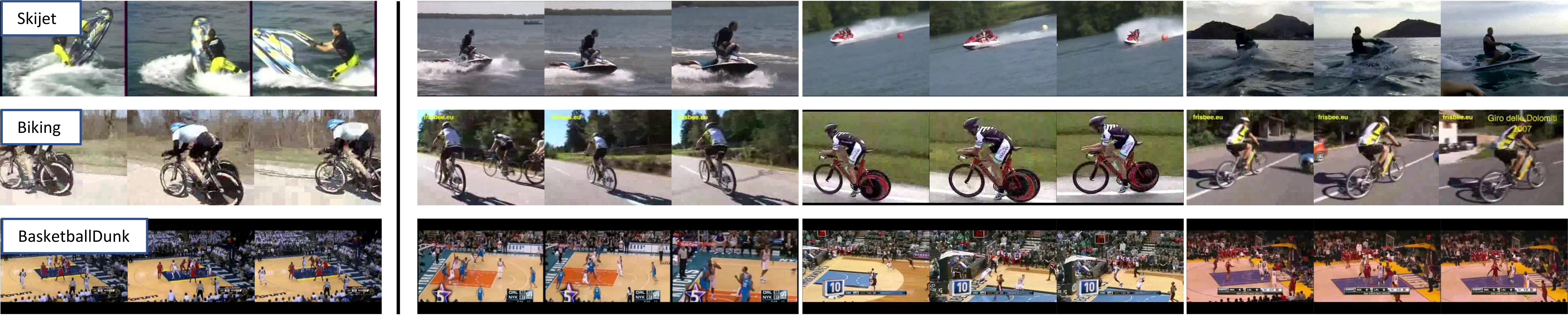}
    \caption{Examples of video retrieval with 3 nearest neighbours.}
    % \vspace{-6pt}
    \label{fig:retriv}
\end{figure}

Finally we present qualitative results using the video retrieval setup, where we pull the nearest neighbours of the query clip with high similarity. We show the input query and the retrieved clips with higher cosine similarity values in Fig. \ref{fig:retriv}. As can be seen, videos with similar semantic appearance and motions are captured, indicating that our method learns important semantically relevant video features.

\section{Conclusions}
We present a contrastive video representation learning method that uses curriculum learning for selecting the positive samples used by the contrastive loss. ConCur starts the contrastive training with easy positive samples which are temporally close and semantically similar, and progressively samples harder positives that are temporally away and semantically dissimilar. The experiments conducted in this paper show that our method improves performance versus blind sampling of positives from the entire input video. We also show that there is an optimal level of difficulty where the performance is maximum. To learn better context-aware representations, we also propose the auxiliary task of predicting the temporal distance between a positive pair of clips. Ablation studies show the effectiveness of each of the components of our method. We achieve state-of-the-art performance with two benchmark datasets on video action recognition and video retrieval using two different encoders. 

\section*{Acknowledgements}
We would like to thank BMO Bank of Montreal and Mitacs for funding this research. We are also thankful to SciNet HPC Consortium for helping with the computing resources.

\bibliographystyle{unsrt}  
\bibliography{ref}

\end{document}